\documentclass[runningheads]{llncs}
\usepackage{graphicx}
\usepackage{amsmath,amssymb} 
\usepackage{color}
\usepackage{enumitem}
\usepackage{amsfonts}
\usepackage{booktabs}
\usepackage{multirow}
\usepackage{subfig}
\usepackage{afterpage}
\usepackage{placeins}

\usepackage{floatrow}

\usepackage[width=122mm,left=12mm,paperwidth=146mm,height=193mm,top=12mm,paperheight=217mm]{geometry}

\begin{document}

\pagestyle{headings}
\mainmatter

\title{Generative Image Modeling using Style and Structure Adversarial Networks} 


\author{Xiaolong Wang, Abhinav Gupta}

\institute{Robotics Institute, Carnegie Mellon University}

\maketitle

\begin{abstract}
Current generative frameworks use end-to-end learning and generate images by sampling from uniform noise distribution. However, these approaches ignore the most basic principle of image formation: images are product of: (a) Structure: the underlying 3D model; (b) Style: the texture mapped onto structure. In this paper, we factorize the image generation process and propose Style and Structure Generative Adversarial Network (${\text{S}^2}$-GAN). Our ${\text{S}^2}$-GAN has two components: the Structure-GAN generates a surface normal map; the Style-GAN takes the surface normal map as input and generates the 2D image. Apart from a real vs. generated loss function, we use an additional loss with computed surface normals from generated images. The two GANs are first trained independently, and then merged together via joint learning.  We show our ${\text{S}^2}$-GAN model is interpretable, generates more realistic images and can be used to learn unsupervised RGBD representations.

\end{abstract}
\section{Introduction}
Unsupervised learning of visual representations is one of the most fundamental problems in computer vision. There are two common approaches for unsupervised learning: (a) using a discriminative framework with auxiliary tasks where supervision comes for free, such as context prediction~\cite{Doersch2014,CarlUnsup2015} or temporal embedding~\cite{WangUnsup2015,nyuUnsup15,ngVideo12,LiEdge16,of_iccv2015,misra2016unsupervised}; (b) using a generative framework where the underlying model is compositional and attempts to generate realistic images~\cite{goodfellow2014generative,Kingma14,Gregor15,Li15}. The underlying hypothesis of the generative framework is that if the model is good enough to generate novel and realistic images, it should be a good representation for vision tasks as well. Most of these generative frameworks use end-to-end learning to generate RGB images from control parameters ($z$ also called noise since it is sampled from a uniform distribution). Recently, some impressive results~\cite{Alec15} have been shown on restrictive domains such as faces and bedrooms.

\begin{figure}[t]
\centering
\includegraphics[width=0.99\textwidth]{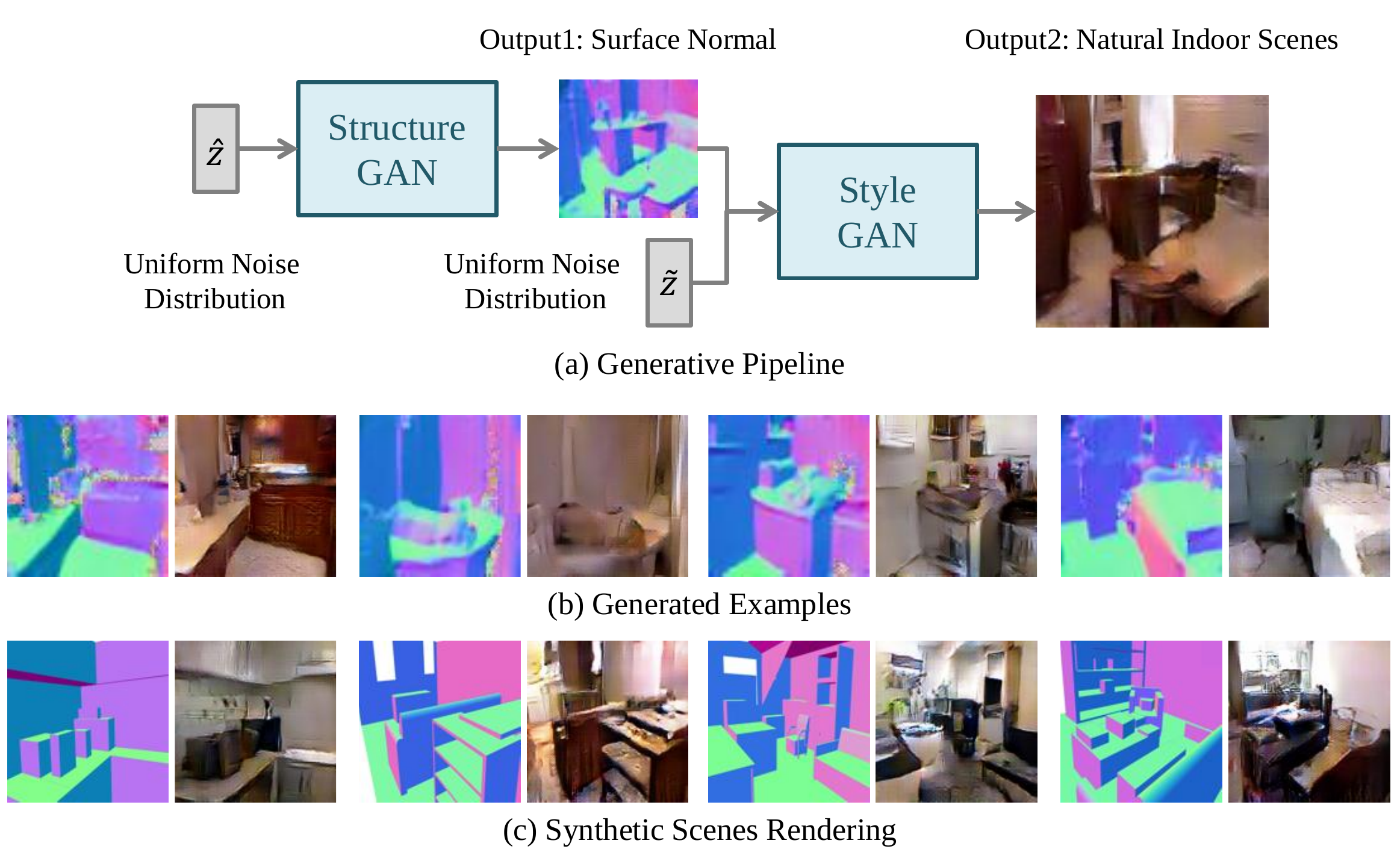}
\footnotesize{\caption{\label{fig:teaser}
\footnotesize{(a) Generative Pipeline: Given $\hat{z}$ sampled from uniform distribution, our Structure-GAN generates a surface normal map as output. This surface normal map is then given as input with $\tilde{z}$ to a second generator network (Style-GAN) and outputs an image. (b) We show examples of generated surface normal maps and images. (c) Our Style-GAN can be used as a rendering engine: given a synthetic scene, we can use it to render a realistic image. To visualize the normals, we represent facing right with blue,  horizontal surface with green, facing left with red (\textcolor{blue}{blue $\rightarrow$ X}; \textcolor{green}{green $\rightarrow$ Y}; \textcolor{red}{red $\rightarrow$ Z}).}}
}
\end{figure}

However, these approaches ignore one of the most basic underlying principles of image formation. Images are a product of two separate phenomena: {\bf Structure:} this encodes the underlying geometry of the scene. It refers to the underlying mesh, voxel representation etc. {\bf Style:} this encodes the texture on the objects and the illumination. In this paper, we build upon this IM101 principle of image formation and factor the generative adversarial network (GAN) into two generative processes as Fig.~\ref{fig:teaser}. The first, a structure generative model (namely Structure-GAN), takes $\hat{z}$ and generates the underlying 3D structure ($y_{3D}$) for the scene. The second, a conditional generative network (namely Style-GAN), takes $y_{3D}$ as input and noise $\tilde{z}$ to generate the image $y_I$. We call this factored generative network Style and Structure Generative Adversarial Network (${\text{S}^2}$-GAN).

{\bf Why ${\text{S}^2}$-GAN?} We believe there are fourfold advantages of factoring the style and structure in the image generation process. Firstly, factoring style and structure simplifies the overall generative process and leads to more realistic high-resolution images. It also leads to a highly stable and robust learning procedure. Secondly, due to the factoring process, ${\text{S}^2}$-GAN is more interpretable as compared to its counterparts. One can even factor the errors and understand where the surface normal generation failed as compared to texture generation. Thirdly, as our results indicate,  ${\text{S}^2}$-GAN allows us to learn RGBD representation in an unsupervised manner. This can be crucial for many robotics and graphics applications. Finally, our Style-GAN can also be thought of as a learned rendering engine which, given any 3D input, allows us to render a corresponding image. It also allows us to build applications where one can modify the underlying 3D structure of an input image and render a completely new image.

However, learning ${\text{S}^2}$-GAN is still not an easy task. To tackle this challenge, we first learn the Style-GAN and Structure-GAN in an independent manner. We use the NYUv2 RGBD dataset~\cite{Silberman12} with more than 200K frames for learning the initial networks. We train a Structure-GAN using the ground truth surface normals from Kinect. Because the perspective distortion of texture is more directly related to normals than to depth, we use surface  normal to represent image structure in this paper. We learn in parallel our Style-GAN which is conditional on the ground truth surface normals. While training the Style-GAN, we have two loss functions: the first loss function takes in an image and the surface normals and tries to predict if they correspond to a real scene or not. However, this loss function alone does not enforce explicit pixel based constraints for aligning generated images with input surface normals.  To enforce the pixel-wise constraints, we make the following assumption: if the generated image is realistic enough, we should be able to reconstruct or predict the 3D structure based on it. We achieve this by adding another discriminator network. More specifically, the generated image is not only forwarded to the discriminator network in GAN but also a input for the trained surface normal predictor network. Once we have trained an initial Style-GAN and Structure-GAN, we combine them together and perform end-to-end learning jointly where images are generated from $\hat{z}, \tilde{z}$ and fed to discriminators for real/fake task.

\section{Related Work}
Unsupervised learning of visual representation is one of the most challenging problems in computer vision. There are two primary approaches to unsupervised learning. The first is the discriminative approach where we use auxiliary tasks such that ground truth can be generated without labeling. Some examples of these auxiliary tasks include predicting: the relative location of two patches~\cite{CarlUnsup2015}, ego-motion in videos~\cite{Agrawal15,Jayaraman15}, physical signals~\cite{Owens16,pintoicra16,pintoeccv2016}.

A more common approach to unsupervised learning is to use a generative framework. Two types of generative frameworks have been used in the past. Non-parametric approaches perform matching of an image or patch with the database for tasks such as texture synthesis~\cite{efros99} or super-resolution~\cite{Freeman02}. In this paper, we are interested in developing a parametric
model of images. One common approach is to learn a low-dimensional representation which can be used to reconstruct an image. Some examples include the deep auto-encoder~\cite{Bengio07,quocle12} or  Restricted Boltzmann machines (RBMs)~\cite{Ranzato10,Osindero2008,hinton06,Honglak09,Taylor06}. However, in most of the above scenarios it is hard to generate new images since sampling in latent space is not an easy task.
The recently proposed Variational auto-encoders (VAE)~\cite{Kingma14,Gregor15} tackles this problem by generating images with variational sampling approach. However, these approaches are restricted to simple datasets such as MNIST. To generate interpretable images with richer information, the VAE is extended to be conditioned on captions~\cite{Mansimov15} and graphics code~\cite{Kulkarni15}. Besides RBMs and auto-encoders, there are also many novel generative models in recent literature~\cite{Dosovitskiy15,TatarchenkoDB15,Theis15,Oord16}. For example, Dosovitskiy et al.~\cite{Dosovitskiy15} proposed to use CNNs to generate chairs.

In this work, we build our model based on the Generative Adversarial Networks (GANs) framework proposed by Goodfellow et al.~\cite{goodfellow2014generative}. This framework was extended by Denton et al.~\cite{Denton15} to generate images. Specifically, they proposed to use a Laplacian pyramid of adversarial networks to generate images in a coarse to fine scheme. However, training these networks is still tricky and unstable. Therefore, an extension DCGAN~\cite{Alec15} proposed good practices for training adversarial networks and demonstrated promising results in generating images. There are more extensions include using conditional variables~\cite{Mirza15,MathieuCL15,RGAN16}. For instance, Mathieu et al.~\cite{MathieuCL15} introduced to predict future video frames conditioned on the previous frames. In this paper, we further simplify the image generation process by factoring out the generation of 3D structure and style.

In order to train our ${\text{S}^2}$-GAN we combine adversarial loss with 3D surface normal prediction loss~\cite{deep3d15,eigen15,Fouhey13a,Ladicky14b} to provide extra constraints during learning. This is also related to the idea of combining multiple losses for better generative modeling~\cite{MakhzaniSJG15,LarsenSW15,DosovitskiyB16}. For example, Makhzani et al.~\cite{MakhzaniSJG15} proposed an adversarial auto-encoder which takes the adversarial loss as an extra constraint for the latent code during training the auto-encoder. Finally, the idea of factorizing image into two separate phenomena has been well studied in~\cite{barrow1978,josh2000,fouhey15,sczhu98}, which motivates us to decompose the generative process to structure and style. We use the RGBD data from NYUv2 to factorize and learn a ${\text{S}^2}$-GAN model.

\section{Background for Generative Adversarial Networks}
The Generative Adversarial Networks (GAN)~\cite{goodfellow2014generative} contains two models: generator $G$ and discriminator $D$. The generator $G$ takes the input which is a latent random vector $z$ sampled from uniform noise distribution and tries to generate a realistic image. The discriminator  $D$ performs binary classification to distinguish whether an image is generated from $G$ or it is a real image. Thus the two models are competing against each other (hence, adversarial): network $G$ will try to generate images which will be hard for $D$ to differentiate from real image, meanwhile network $D$ will learn to avoid getting fooled by $G$.

Formally, we optimize the networks using gradient descent with batch size $M$.  We are given samples as $\textbf{X} = (X_1, ..., X_M)$ and a set of $z$ sampled from uniform distribution as $\textbf{Z} = (z_1,...,z_M)$. The training of GAN is an iterative procedure with 2 steps: (i) fix the parameters of network $G$ and optimize network $D$; (ii) fix network $D$ and optimize network $G$. The loss for training network $D$ is,
{\footnotesize
\begin{eqnarray}\label{eq:loss_d}
L^{D}(\textbf{X},\textbf{Z}) = \sum_{i=1}^{M/2} L(D(X_i), 1) + \sum_{i=M/2 + 1}^{M} L(D(G(z_i)),0).
\end{eqnarray}
}
Inside a batch, half of images are real and the rest $G(z_i)$ are images generated by $G$ given $z_i$. $D(X_i) \in [0,1]$ represents the binary classification score given input image $X_i$. $L(y^*, y) = -[y \log(y^*) + (1-y)log(1-y^*)]$ is the binary entropy loss. Thus the loss Eq.~\ref{eq:loss_d} for network $D$ is optimized to classify the real image as label $1$ and the generated image as $0$. On the other hand, the generator $G$ is trying to fool $D$ to classify the generated image as a real image via minimizing the loss:
{\footnotesize
\begin{eqnarray}\label{eq:loss_g}
L^{G}(\textbf{Z}) = \sum_{i=M/2 + 1}^{M} L(D(G(z_i)),1).
\end{eqnarray}
}

\vspace{-0.5in}
\section{Style and Structure GAN}
GAN and DCGAN approaches directly generate images from the sampled $z$. Instead, we use the fact that image generation has two components: (a) generating the underlying structure based on the objects in the scene; (b) generating the texture/style on top of this 3D structure. We use this simple observation to decompose the generative process into two procedures: (i) Structure-GAN - this process generates surface normals from sampled $\hat{z}$ and (ii) Style-GAN - this model generates the images taking as input the surface normals and another latent variable $\tilde{z}$ sampled from uniform distribution. We train both models with RGBD data, and the ground truth surface normals are obtained from the depth.

\begin{table}[t]
   \begin{center}
   \setlength{\tabcolsep}{0.05cm}
   \subfloat {
  \scriptsize{
  \begin{tabular}{|l|cccccccccc|}
      \hline
      \scriptsize{Structure-GAN(G)}   & fc  & uconv    & conv & conv & conv & conv & uconv & conv & uconv & conv\\ \hline
      \scriptsize{Input Size} & $-$  & $9$    & $18$    & $18$ & $18$  & $18$ & $18$ & $36$ & $36$ & $72$\\
      \scriptsize{Kernel Number}  & $ 9 \times 9 \times 64$  & $128$  & $128$ & $256$ & $512$  & $512$ & $256$  & $128$ & $64$ & $3$\\
      \scriptsize{Kernel Size}    & $-$   & $4$  & $3$   & $3$ & $3$ & $3$ & $4$ & $3$ & $4$ & $5$\\
      \scriptsize{Stride}    & $-$  & $2(up)$  & $1$ & $1$ & $1$ & $1$ & $2(up)$ & $1$ & $2(up)$ & $1$\\
      \hline
    \end{tabular}}
    }

    \subfloat {
  \scriptsize{
  \begin{tabular}{|l|cccccc|}
      \hline
      \scriptsize{Structure-GAN(D)}   & conv & conv    & conv & conv & conv & fc \\ \hline
      \scriptsize{Input Size} & $72$  & $36$    & $36$    & $18$ & $9$  & $-$ \\
      \scriptsize{Kernel Number}  & $64$  & $128$  & $256$ & $512$ & $128$  & $1$ \\
      \scriptsize{Kernel Size}    & $5$   & $5$  & $3$   & $3$ & $3$ & $-$ \\
      \scriptsize{Stride}    & $2$  & $1$  & $2$ & $2$ & $1$ & $-$ \\
      \hline
    \end{tabular}}
    }
  \subfloat {
  \scriptsize{
  \begin{tabular}{|l|cccccc|}
      \hline
      \scriptsize{Style-GAN(D)}   & conv & conv    & conv & conv & conv & fc \\ \hline
      \scriptsize{Input Size} & $128$  & $64$    & $32$    & $16$ & $8$  & $-$ \\
      \scriptsize{Kernel Number}  & $64$  & $128$  & $256$ & $512$ & $128$  & $1$ \\
      \scriptsize{Kernel Size}    & $5$   & $5$  & $3$   & $3$ & $3$ & $-$ \\
      \scriptsize{Stride}    & $2$  & $2$  & $2$ & $2$ & $1$ & $-$ \\
      \hline
    \end{tabular}}
    }
  \end{center}
  \footnotesize{\caption{
  \footnotesize{Network architectures. Top: generator of Structure-GAN; bottom: discriminator of Structure-GAN (left) and discriminator of Style-GAN (right). ``conv'' means convolutional layer, ``uconv'' means fractionally-strided convolutional (deconvolutional) layer, where $2(up)$ stride indicates 2x resolution. ``fc'' means fully connected layer.}  }
  \label{tbl:arch}}

\end{table}

\subsection{Structure-GAN}
\begin{figure}[t]
\center
\includegraphics[width=\textwidth]{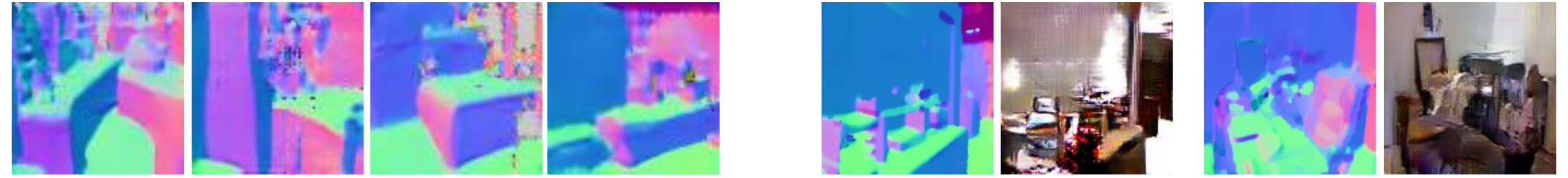}
\footnotesize{\caption{\label{fig:example}
\footnotesize{Left: 4 Generated Surface Normal maps. Right: 2 Pairs of rendering results on ground truth surface normal maps using the Style-GAN without pixel-wise constraints. }}
}
\end{figure}

We can directly apply GAN framework to learn how to generate surface normal maps. The input to the network $G$ will be $\hat{z}$ sampled from uniform distribution and the output is a surface normal map. We use a 100-d vector to represent the $\hat{z}$ and the output is in size of $72\times72\times3$ (Fig.~\ref{fig:example}). The discriminator $D$ will learn to classify the generated surface normal maps from the real maps obtained from depth. We introduce our network architecture as following.

{\bf{Generator network}}. As Table~\ref{tbl:arch} (top row) illustrates, we apply a 10-layer model for the generator. Given a $100$-d $\hat{z}$ as input, it is first fully connected to a 3D block ($9 \times 9 \times 64$). Then we further perform convolutional operations on top of it and generate the surface normal map in the end. Note that ``uconv'' represents fractionally-strided convolution~\cite{Alec15}, which is also called as deconvolution. We follow the settings in~\cite{Alec15} and use Batch Normalization~\cite{BN15} and ReLU activations after each layer except for the last layer, where a TanH activation is applied.

{\bf{Discriminator network}}. We show the 6-layer network architecture in   Table~\ref{tbl:arch} (bottom left). Taking  an image as input, the network outputs a single number which predicts the input surface normal is real or generated. We use LeakyReLU~\cite{Maas13,xu15} for activation functions as in~\cite{Alec15}. However, we do not apply Batch Normalization here. In our case, we find that the discriminator network easily finds trivial solutions with Batch Normalization.

\subsection{Style-GAN}

Given the RGB images and surface normal maps from Kinect, we train another GAN in parallel to generate images conditioned on surface normals. We call this network Style-GAN. First, we modify our generator network to a conditional GAN as proposed in~\cite{Mirza15,Denton15}. The conditional information, i.e., surface normal maps, are given as additional inputs for both the generator $G$ and the  discriminator $D$. Augmenting surface normals as an additional input to $D$ not only forces the generated image to look real, but also implicitly enforces the generated image to match the surface normal map. While training this discriminator, we only consider real RGB images and their corresponding surface normals as the positive examples. Given more cues from surface normals, we  generate higher resolution of $128 \times 128 \times 3$ images with the Style-GAN.

Formally, we have a batch of RGB images $\textbf{X} = (X_1, ..., X_M)$ and their corresponding surface normal maps $\textbf{C} = (C_1,...,C_M)$, as well as samples from noise distribution ${\tilde{\textbf{Z}}} = (\tilde{z}_1, ..., \tilde{z}_M)$. We reformulate the generative function from $G(\tilde{z}_i)$ to $G( C_i, \tilde{z}_i)$ and discriminative function is changed from $D(X_i)$ to $D( C_i, X_i)$. Then the loss of discriminator network in Eq.~\ref{eq:loss_d} can be reformulated as,
{\footnotesize
\begin{eqnarray}\label{eq:loss_d2}
L^{D}_{cond}(\textbf{X},\textbf{C},\tilde{\textbf{Z}}) = \sum_{i=1}^{M/2} L(D(C_i,X_i), 1) + \sum_{i=M/2 + 1}^{M} L(D(C_i, G(C_i,\tilde{z}_i)),0),
\end{eqnarray}
}
and the loss of generator network in Eq.~\ref{eq:loss_g} can be reformulated as,
{\footnotesize
\begin{eqnarray}\label{eq:loss_g2}
L^{G}_{cond}(\textbf{C},\tilde{\textbf{Z}}) = \sum_{i=M/2 + 1}^{M}  L(D(C_i, G(C_i,\tilde{z}_i)),1).
\end{eqnarray}
}
We apply the same scheme of iterative training. By doing this, we can generate the images with network $G$ as visualized in Fig.~\ref{fig:example} (right).

\begin{figure}[t]
\center
\includegraphics[width=0.95\textwidth]{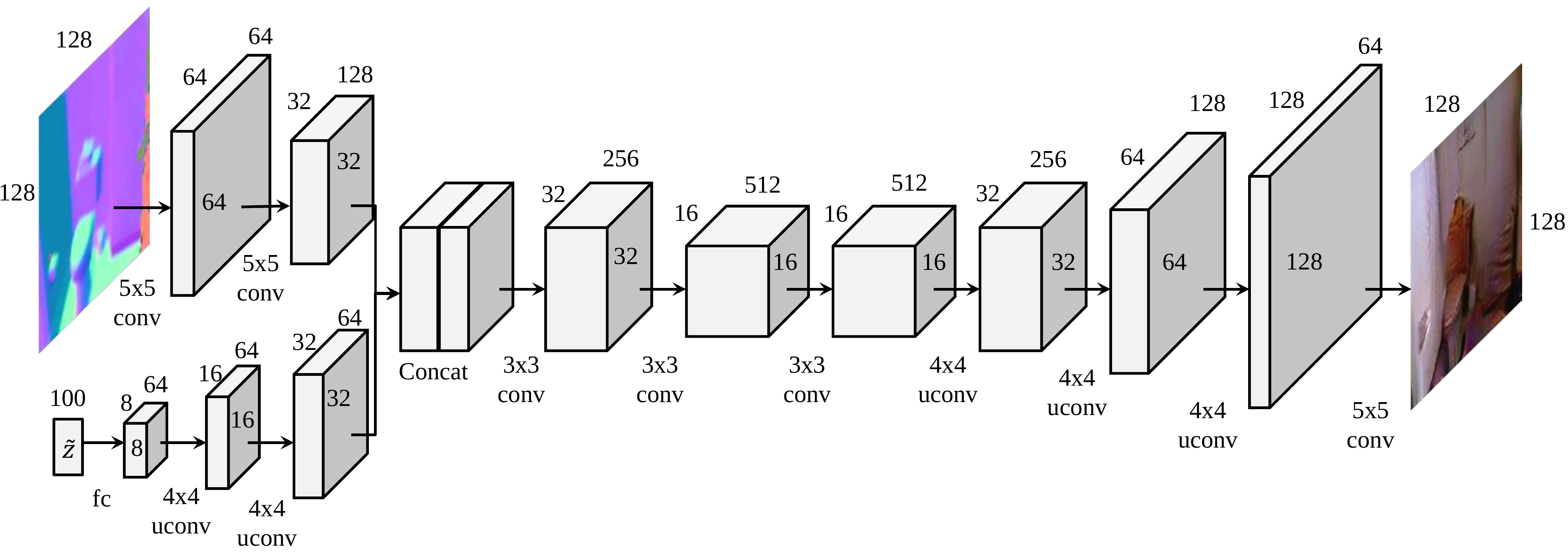}
\footnotesize{\caption{\label{fig:G2_arch}\footnotesize{ The architecture of the generator in Style-GAN.
}}
}

\end{figure}

\noindent {\bf Network architecture.} We show our generator as Fig.~\ref{fig:G2_arch}. Given a $128 \times 128 \times 3$ surface normal map and  a 100-d $\tilde{z}$ as input, they are firstly forwarded to convolutional and deconvolutional layers respectively and then concatenated to form $32 \times 32 \times 192$ feature maps. On top of these feature maps, 7 layers of convolutions and deconvolutions are further performed. The output of the network is a $128 \times 128 \times 3$ RGB image. For the discriminator, we apply the similar architecture of the one in Structure-GAN (bottom right in Table.~\ref{tbl:arch}). The input for the network is the concatenation of surface normals and images ($128 \times 128 \times 6$).

\subsection{Multi-task Learning with Pixel-wise Constraints}
The Style-GAN can make the generated image look real and also enforce  it to match the provided surface normal maps implicitly. However, as shown Fig.~\ref{fig:example}, the images are noisy and the edges are not well aligned with the edges in the surface normal maps. Thus, we propose to add a pixel-wise constraint to explicitly guide the generator to align the outputs with the input surface normal maps.

\begin{figure}[t]
\center
\includegraphics[width=0.9\textwidth]{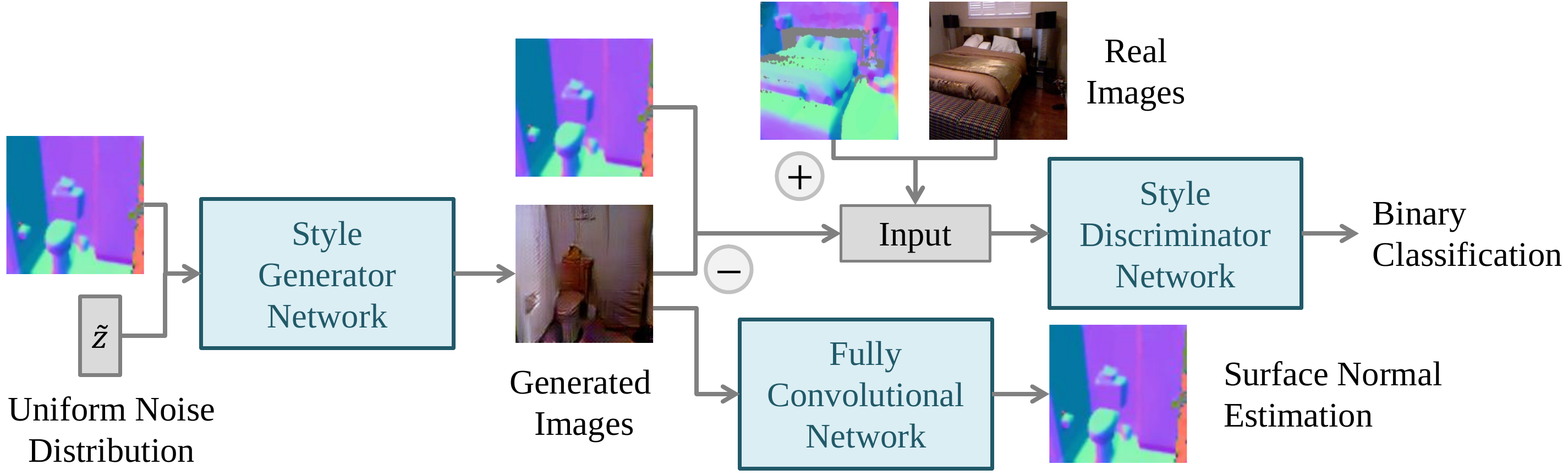}
\footnotesize{
\caption{\label{fig:model2}
\footnotesize{
Our Style-GAN.  Given the ground truth surface normals and $\tilde{z}$ as inputs, the generator $G$ learns to generate RGB images. The supervision comes from two networks: The discriminator network takes the generated images, real images and their corresponding normal maps as inputs to perform classification; The FCN takes the generated images as inputs and predict the surface normal maps.
}}
}
\end{figure}

We make the following assumption: If the generated image is real enough, it can be used for reconstructing the surface normal maps. To encode this constraint, we train another network for surface normal estimation. We modify the Fully Convolutional Network (FCN)~\cite{Long15} with the classification loss as mentioned in~\cite{deep3d15} for this task. More specifically, we quantize the surface normals to $40$ classes with k-means clustering as in~\cite{deep3d15,Ladicky14} and the loss is defined as

{\footnotesize
\begin{eqnarray}\label{eq:loss_fcn}
L^{FCN}(\textbf{X},\textbf{C}) = \frac{1}{K \times K} \sum_{i=1}^{M} \sum_{k=1}^{K \times K} L_s (F_k(X_i), C_{i,k}),
\end{eqnarray}
}
where $L_s$ means the softmax loss and the output surface normal map is in $K \times K$ dimension, and $K = 128$ is in the same size of input image. $F_k(X_i)$ is the output of $k$th pixel in the $i$th sample. $C_{i,k} (1 \leqslant C_{i,k} \leqslant 40)$  is the label for the $k$th pixel in sample $i$. Thus the loss is designed to enforce each pixel in the image to generate accurate surface normal. Note that when training the FCN, we use the RGBD data which provides indoor scene images and ground truth surface normals. The model is trained from scratch without ImageNet pre-training.

{\bf FCN architecture}. We apply the AlexNet~\cite{Krizhevsky12} following the same training scheme as~\cite{Long15}, with modifications on the last 3 layers. Given a generated $128 \times 128$ image, it is first upsampled to $512 \times 512$ before feeding into the FCN. For the two layers before the last layer, we use smaller kernel numbers of $1024$ and $512$. The last layer is a deconvolutional layer with stride 2. In the end, upsampling (4x resolution) is further applied to generate the high quality results.

Given the trained FCN model, we can use it as an additional supervision (constraint) in the adversarial learning. Our final model is illustrated in Fig.~\ref{fig:model2}. During training, not only the gradients from the classification loss of $D$ will be passed down to $G$, but also the surface normal estimation loss from the FCN is passed through the generated image to $G$. This way, the adversarial loss from  $D$ will make the generated images look real, and the FCN will give pixel-wise constraints to make the generated images aligned with surface normal maps.

Formally, we combine the two losses in Eq.~\ref{eq:loss_g2} and Eq.~\ref{eq:loss_fcn} for the generator $G$,
{\footnotesize
\begin{eqnarray}\label{eq:loss_g3}
L^{G}_{multi}(\textbf{C},\tilde{\textbf{Z}}) = L^{G}_{cond}(\textbf{C},\tilde{\textbf{Z}}) +  L^{FCN}(G(\textbf{C},\tilde{\textbf{Z}}),\textbf{C}),
\end{eqnarray}
}
where $G(\textbf{C},\tilde{\textbf{Z}})$ represents the generated images given a batch of surface normal maps $\textbf{C}$ and noise $\tilde{\textbf{Z}}$. The training procedure for this model is similar to the original adversarial learning, which includes three steps in each iteration:
\begin{itemize}[noitemsep]
\item  Fix the generator $G$, optimize the discriminator $D$ with Eq.~\ref{eq:loss_d2}.
\item  Fix the FCN and the discriminator $D$, optimize the generator $G$ with Eq.~\ref{eq:loss_g3}.
\item Fix the generator $G$, fine-tune FCN using generated and real images.
\end{itemize}
Note that the parameters of FCN model are fixed in the beginning of multi-task learning, i.e., we do not fine-tune FCN in the beginning. The reason is the generated images are not good in the beginning, so feeding bad examples to FCN seems to make the surface normal prediction worse.

\begin{figure}[t]
\center
\includegraphics[width=0.95\textwidth]{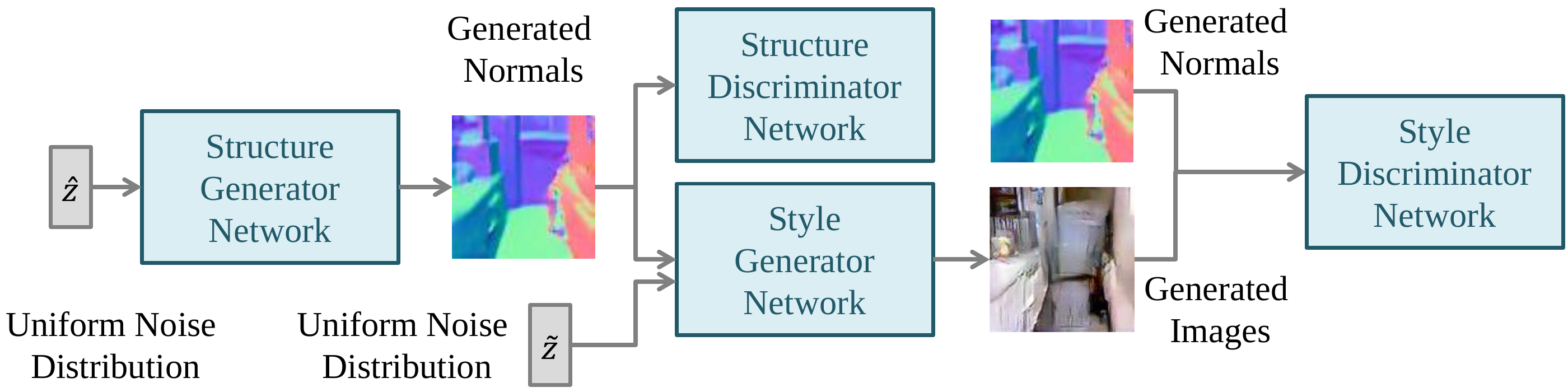}
\footnotesize{
\caption{\label{fig:model_joint}
\footnotesize{Full model of our ${\text{S}^2}$-GAN. It can directly generate RGB images given $\hat{z},\tilde{z}$ as inputs. For simplicity, we do not visualize the positive samples in training. During joint learning, the loss from Style-GAN is also passed down to the Structure-GAN.
}}
}
\end{figure}

\subsection{Joint Learning for ${\text{S}^2}$-GAN }
After training the Structure-GAN and Style-GAN independently, we merge all networks and train them jointly. As Fig.~\ref{fig:model_joint} shows, our full model includes surface normal generation from Structure-GAN, and based on it the Style-GAN generates the image. Note that the generated normal maps are first passed through an upsampling layer with bilinear interpolation before they are forwarded to the Style-GAN. Since we do not use ground truth surface normal maps to generate the images, we remove the FCN constraint from the Style-GAN. The discriminator in Style-GAN takes generated normals and images as negative samples, and ground truth normals and real images as positive samples.

For the Structure-GAN, the generator network  receives not only the gradients from the discriminator of Structure-GAN, but also the gradients passed through the generator of Style-GAN. In this way, the network is forced to generate surface normals which not only are realistic but also help generate better RGB images. Formally, the loss for the generator network of Structure-GAN can be represented as combining Eq.~\ref{eq:loss_g} and Eq.~\ref{eq:loss_g2},
{\footnotesize
\begin{eqnarray}\label{eq:loss_g_joint}
L^{G}_{joint}(\hat{\textbf{Z}}, \tilde{\textbf{Z}})    = L^{G}(\hat{\textbf{Z}}) + \lambda \cdot L^{G}_{cond}(G(\hat{\textbf{Z}}),\tilde{\textbf{Z}})
\end{eqnarray}
}
where $\hat{\textbf{Z}} = (\hat{z}_1, ..., \hat{z}_M)$ and  ${\tilde{\textbf{Z}}} = (\tilde{z}_1, ..., \tilde{z}_M)$  represent two sets of samples drawn from uniform distribution for Structure-GAN and Style-GAN respectively. The first term in Eq.~\ref{eq:loss_g_joint} represents the adversarial loss from the discriminator of Structure-GAN and the second term represents that the loss of the Style-GAN is also passed down. We set the coefficient $\lambda = 0.1$ and smaller learning rate for Structure-GAN than Style-GAN in the experiments,  so that we can prevent the generated normals from over fitting to the task of generating RGB images via Style-GAN. In our experiments, we find that without  constraining $\lambda$ and learning rates, the loss $L^{G}(\hat{\textbf{Z}})$ easily diverges to high values and the Structure-GAN can no longer generate reasonable surface normal maps.

\begin{figure}[t]
\center
\includegraphics[width=0.99\textwidth]{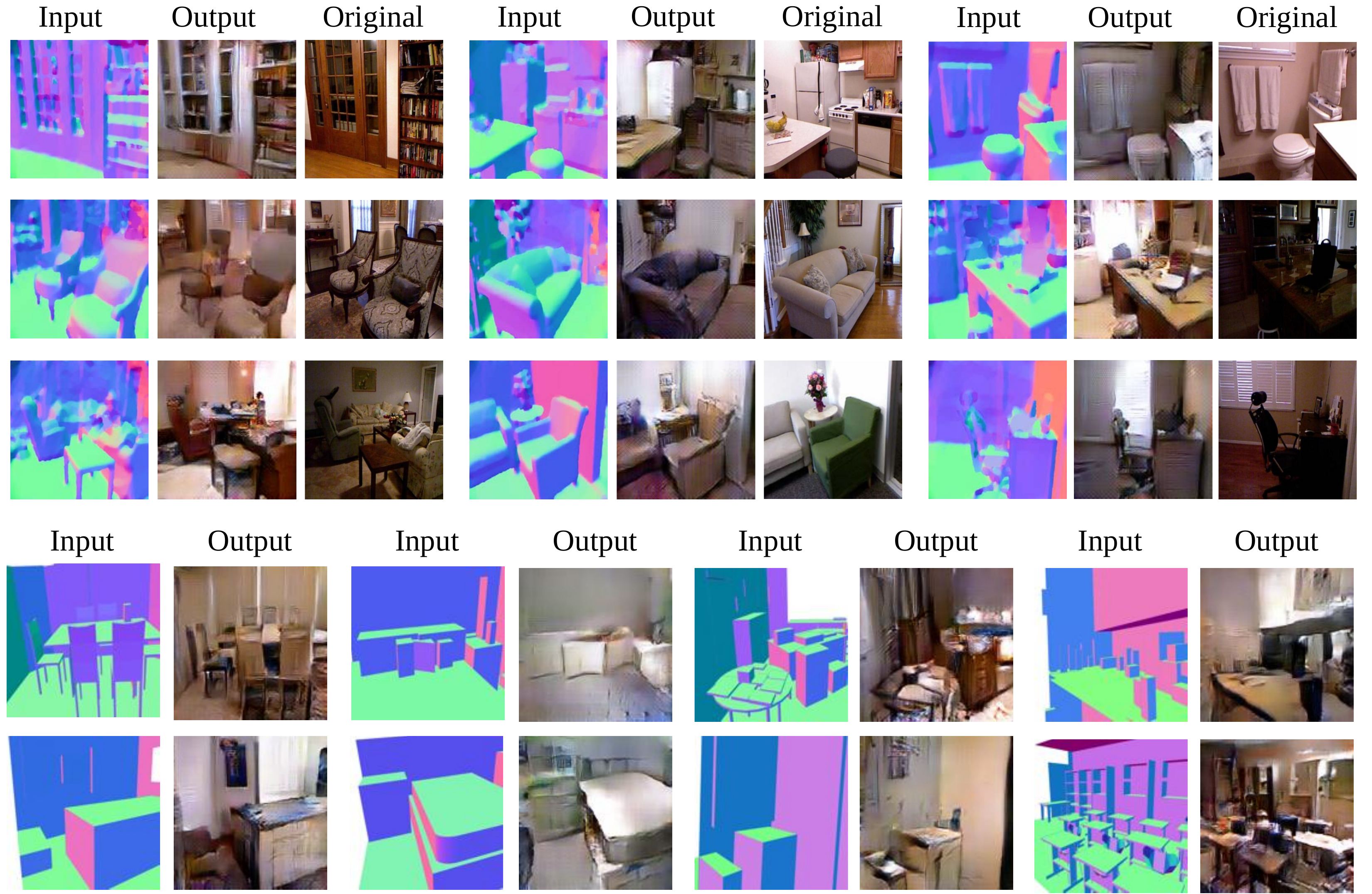}
\footnotesize{ \caption{\label{fig:result} \footnotesize{
Results of Style-GAN conditioned on ground truth surface normals (first 3 rows) and synthetic scenes (last 2 rows). For ground truth normals, we show the input normals, our generated images and the original corresponding images.}}
}
\end{figure}
\begin{figure}[!tbp]
\center
\includegraphics[width=0.95\textwidth]{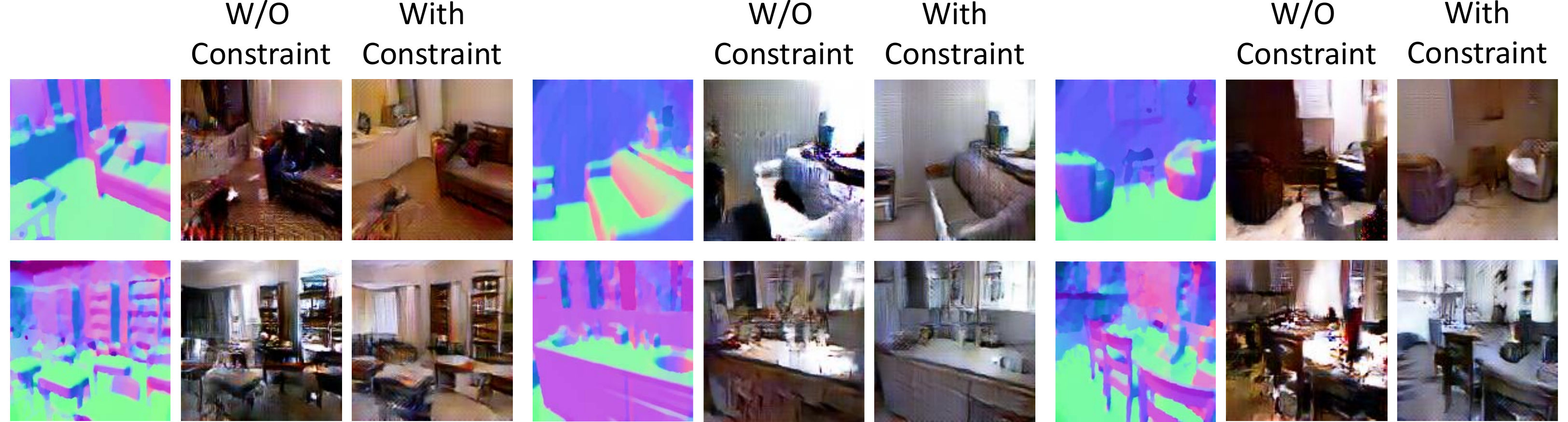}
\footnotesize{
\caption{\label{fig:cmp} \footnotesize{Comparison between models with and without pixel-wise constraints.} }
}
\end{figure}

\begin{figure}[t]
\center
\includegraphics[width=\textwidth]{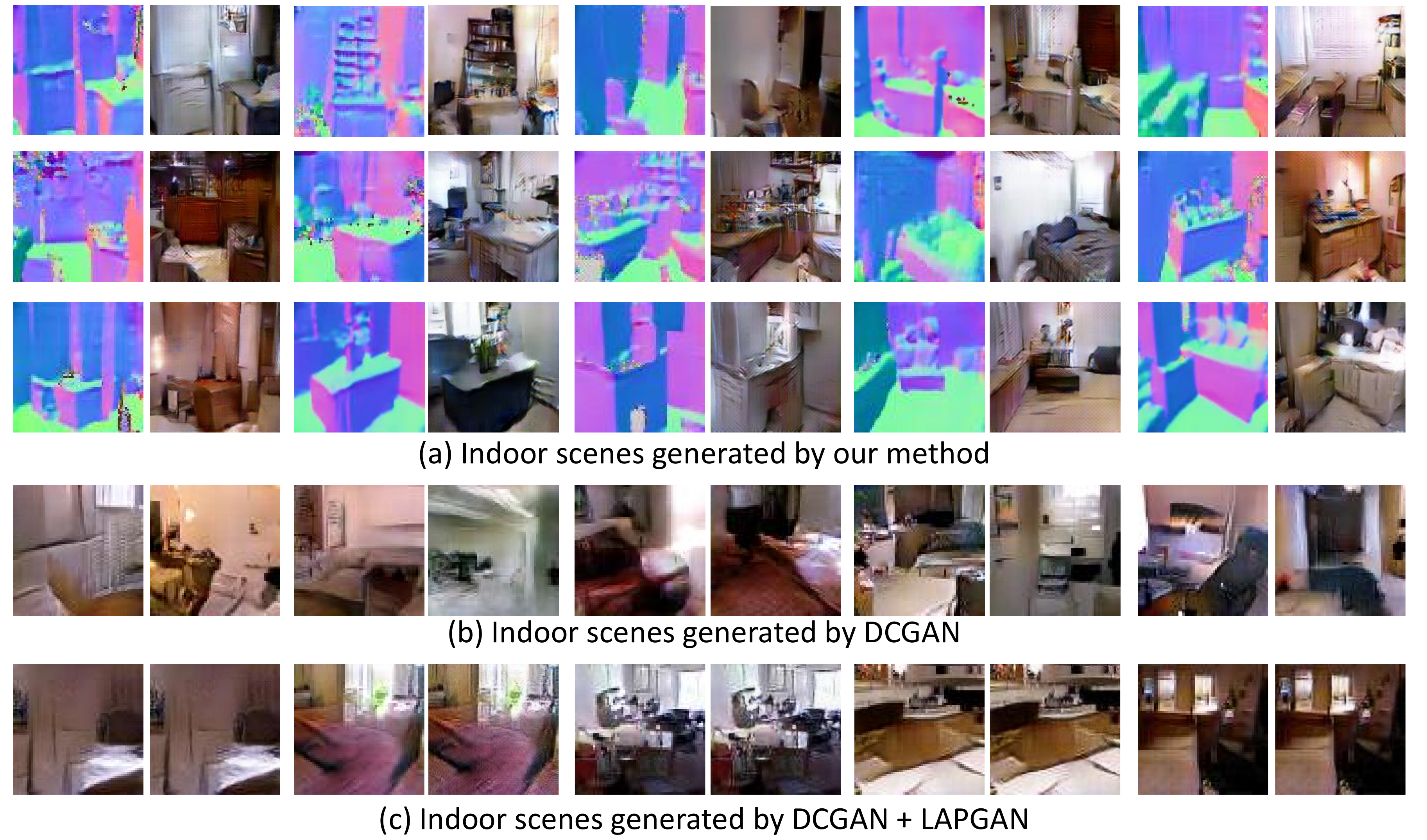}
\footnotesize{
\caption{\label{fig:full} \footnotesize{(a) Pairs of surface normals and images generated by ${\text{S}^2}$-GAN. (b) Results of DCGAN. (c) Results of DCGAN+LAPGAN. For each pair, result on the left is from DCGAN and on the right is applying LAPGAN after it. }}
}
\end{figure}

\section{Experiments}
We perform two types of experiments: (a) We qualitatively and quantitatively evaluate the quality of images generates using our model; (b) We evaluate the quality of unsupervised representation learning by applying the network for different tasks such as image classification and object detection.

\noindent \textbf{Dataset.} We use the NYUv2 dataset~\cite{Silberman12} in our experiment.  We use the raw video data during training and extract 200K frames from the 249 training video scenes. We compute the surface normals from the depth as~\cite{Ladicky14b,deep3d15}.

\noindent \textbf{Parameter Settings.} We follow the parameters in~\cite{Alec15} for training. We trained the models using  Adam optimizer~\cite{Kingma14adam} with momentum term $\beta_1 = 0.5, \beta_2 = 0.999$ and batch size $M=128$. The inputs and outputs for all networks are scaled to $[-1,1]$ (including surface normals and RGB images). During training the Style and Structure GANs separately, we set the learning rate to $0.0002$. We train the Structure-GAN for 25 epochs. For Style-GAN, we first fix the FCN model and train it for 25 epochs, then the FCN model are fine-tuned together with 5 more epochs. For joint learning, we set learning rate as $10^{-6}$ for Style-GAN and $10^{-7}$ for Structure-GAN and train them for 5 epochs.

\noindent \textbf{Baselines.} We have 4 baseline models trained on NYUv2 training set: (a) DCGAN~\cite{Alec15}: it takes uniform noise as input and generate $64 \times 64$ images; (b) DCGAN+LAPGAN: we train a LAPGAN~\cite{Denton15} on top of DCGAN, which takes lower resolution images as inputs and generates $128 \times 128$ images. We apply the same architecture as our Style-GAN for LAPGAN (Fig.~\ref{fig:G2_arch} and Table.~\ref{tbl:arch}). (c) DCGANv2: we train a DCGAN with the same architecture as our Structure-GAN (Table.~\ref{tbl:arch}). (d) DCGANv2+LAPGAN: we train another LAPGAN on top of DCGANv2 as (b) with the same architecture. Note that baseline (d) has the same model complexity as our model.

\begin{figure*}[t]
\center
\includegraphics[width=0.99\textwidth]{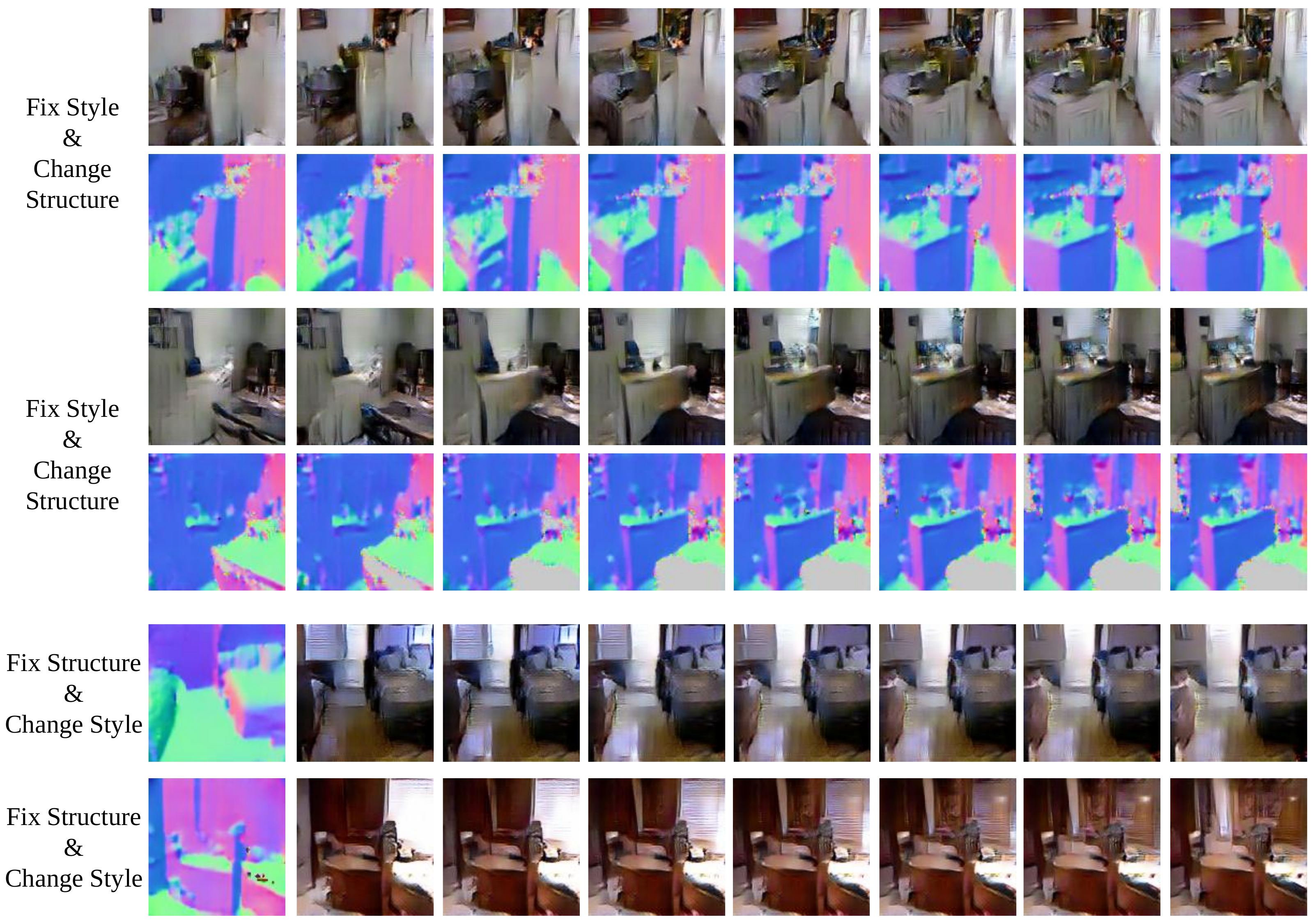}
\footnotesize{
\caption{\label{fig:latent}
\footnotesize{Walking the latent space: Our latent space is more interpretable and we obtain smooth transitions of generated results by interpolating the inputs.}}
}
\end{figure*}

\subsection{Qualitative Results for Image Generation }
\noindent \textbf{Style-GAN Visualization}. Before showing the image generation results of the full ${\text{S}^2}$-GAN model, we first visualize the results of our Style-GAN given the ground truth surface normals on the NYUv2 test set. As illustrated in the first 3 rows of Fig.~\ref{fig:result}, we can generate nice rendering results which are well aligned with the surface normal inputs. By comparing with the original RGB images, we show that our method can generate a different style (illumination, color, texture) of image with the same structure.  We also make comparisons on the results of Style-GAN with/without pixel-wise constraints as visualized in Fig.~\ref{fig:cmp}. We show that if we train the model without the pixel-wise constraint, the output is less smooth and noisier than our approach.

\noindent \textbf{Rendering on Synthetic Scenes}. One application of our Style-GAN is rendering synthetic scenes. We use the 3D models annotated in~\cite{Guo13} to generate the synthetic scenes. We use the scenes corresponding to the NYUv2 test set and make some modifications by rotation, zooming in/out. As the last two rows of Fig.~\ref{fig:result} show, we can obtain very realistic rendering results on 3D models.

\noindent \textbf{${\text{S}^2}$-GAN Visualization}. We now show the results of our full generative model. Given the noise $\hat{z}, \tilde{z}$, our model generate both surface normal maps ($72 \times 72$) and RGB images ($128 \times 128$) after that, as shown in Fig.~\ref{fig:full}(a). We compare with the baselines including DCGAN(Fig.~\ref{fig:full}(b)) and DCGAN+LAPGAN (Fig.~\ref{fig:full}(c)). We can see that our method can generate more structured indoor scenes, i.e.,  it is easier to figure out the structure and objects in our image. We also find that using LAPGAN does not help much improving the qualitative results.

\noindent \textbf{Walking the latent space.} One big advantage of our model is that it is interpretable. Recall that we have two random uniform vectors $\hat{z}, \tilde{z}$ as inputs for Structure and Style networks. We conduct two experiments here:  (i) Fix $\tilde{z}$ (style) and manipulate the structure of images by changing $\hat{z}$; (ii) Fix $\hat{z}$ (structure) and manipulate the style of images by changing $\tilde{z}$. Specifically, given an initial set of $\hat{z}$ and $\tilde{z}$, we pick up a series of 10 random points in $\hat{z}$ or $\tilde{z}$ and gradually add $0.1$ to these points for $6-7$ times. We show that we can obtain smooth transitions in the outputs by interpolating the inputs as Fig.~\ref{fig:latent}. For the example in the first two rows of Fig.~\ref{fig:latent}, we show that by interpolating $\hat{z}$, we can gradually ``grow'' a 3D cube in the room and the style of the RGB images are consistent since we fix the $\tilde{z}$. For the last rows in Fig.~\ref{fig:latent}, we fix the structure of the image and interpolate the $\tilde{z}$ so that the window of the room is gradually shut down.

\noindent \textbf{User study.} We collect 1000 pairs of images randomly generated by our method and DCGAN. We let the  AMT workers to judge which one is more realistic in each pair and $71\%$ of the time they think our approach generates better images.

\begin{figure}[t]
\center
\includegraphics[width=0.95\textwidth]{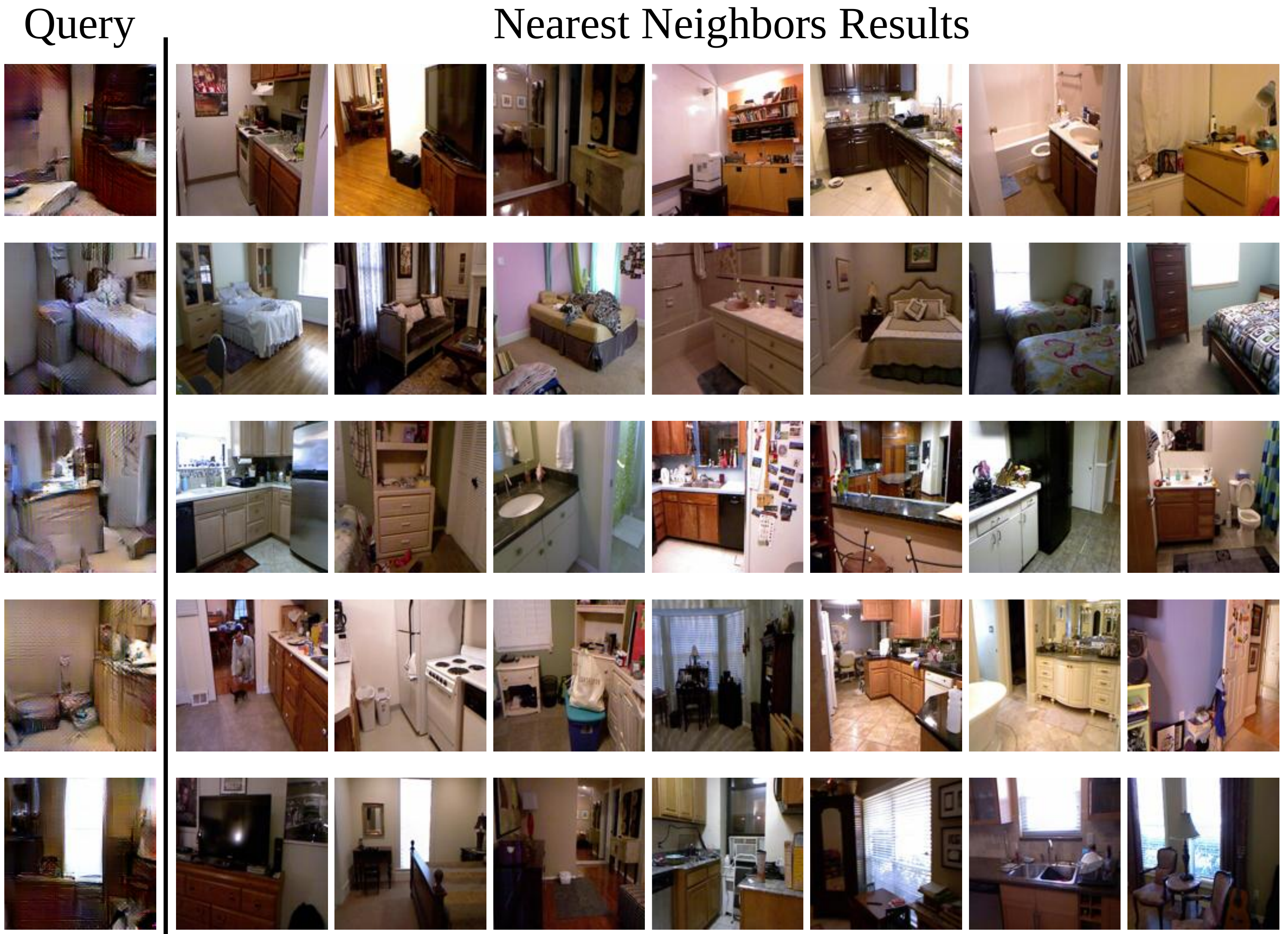}
\footnotesize{
\caption{\label{fig:nntest}
\footnotesize{Nearest neighbors test on generated images.}
}}
\end{figure}

\noindent \textbf{Nearest Neighbors Test.}  To estimate the novelness of our generated images, we apply nearest neighbors test on them. We apply the AlexNet pre-trained on the Places dataset~\cite{Zhou14} as feature extractor. We extract the Pool5 feature of the generated images as well as the real images (both training and testing) from the dataset. We show the results as Fig.~\ref{fig:nntest}. In each row, the first image is generated by our model, which is used as a query. We show the top 7 retrieved real images. We observe that while the images are semantically related, they have different style and structure as compared to nearest neighbors.

\subsection{Quantitative Results for Image Generation}
To evaluate the generated images quantitatively, we apply the AlexNet pre-trained (supervised) on Places~\cite{Zhou14} and ImageNet dataset~\cite{ILSVRC15} to perform classification and detection on them. The motivation is: If the generated images are realistic enough, state of the art classifiers and detectors should fire on them with high scores. We compare our method with the three baselines mentioned in the beginning of experiment: DCGAN, DCGANv2 and DCGANv2+LAPGAN. We generate 10K images for each model and perform evaluation on them.

\begin{figure}[t]
\center
\includegraphics[width=0.95\textwidth]{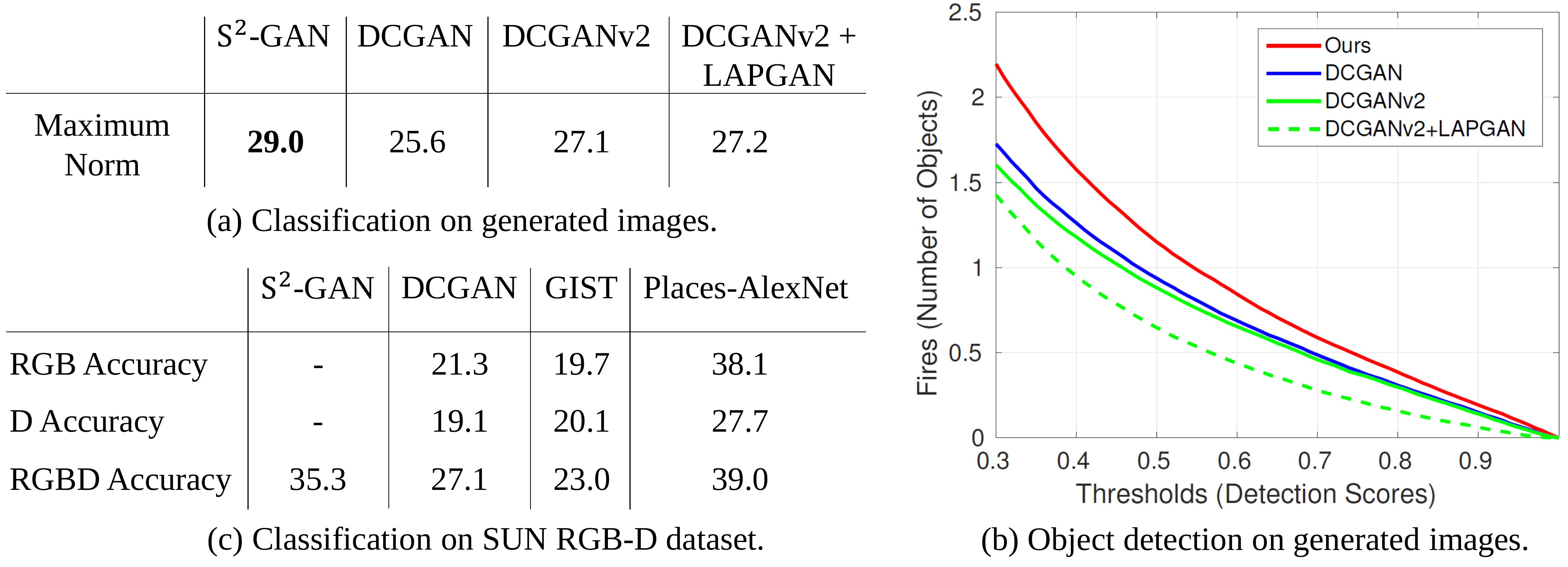}
\footnotesize{\caption{\label{fig:test} \footnotesize{ (a) The Maximum Norm of classification results on generated images. (b) Number of fires over different thresholds for object detection on generated images. (c) Scene classification on SUN RGB-D with our model and other methods (no fine-tuning).}}
}
\end{figure}

\textbf{Classification on generated images.} We apply the Places-AlexNet~\cite{Zhou14} to perform classification on the generated images. If the image is real enough, the Places-AlexNet will give high response in one class during classification. Thus, we can use the maximum norm $||\cdot||_{\infty}$ of the softmax output (i.e., the maximum probability) of Places-AlexNet to represent the image quality. We compute the results for this metric on all generated images and show the mean for different models as Fig.~\ref{fig:test}(a). ${\text{S}^2}$-GAN is around $2\%$ better than the baselines.

\textbf{Object detection on generated images.} We used Fast-RCNN detector~\cite{girshickICCV15fastrcnn} fine-tuned on the NYUv2 dataset with ImageNet pre-trained AlexNet. We then apply the detector on generated images. If the image is realistic enough, the detector should find objects (door, bed, sofa, table, counter etc). Thus, we want to investigate on which images the detector can find more foreground objects.  We plot the curves shown in Fig.~\ref{fig:test}(b) (the x-axis represents the detection threshold, and the y-axis represents average number of detections). We show that the detector can find more foreground objects in the images generated by ${\text{S}^2}$-GAN. At $0.3$ threshold, there are on average $2.2$ detections per image and $1.72$ detections on images generated by DCGAN.

\subsection{Representation Learning for Recognition Tasks}
We now explore whether the representation learned by the discriminator network in our Style-GAN can be transferred to tasks such as scene classification and object detection. Since the input for the network is RGB image and surface normal map, our model can be applied to recognition tasks in RGBD data. We perform the experiments on scene classification on SUN RGB-D dataset~\cite{Song15,Janoch11,Xiao13,Silberman12} as well as object detection on NYUv2 dataset.

\textbf{Scene Classification.} We use the standard train/test split for scene classification in SUN RGB-D dataset, which includes 19 classes with 4852 training and 4660 testing images. We use our model, taking RGB images and normals as inputs, to extract the feature of the second-to-last layer and train SVM on top of it. We compare our method with the  discriminator network in DCGAN and the baselines reported in~\cite{Song15}: GIST~\cite{GIST11} feature as well as Places-AlexNet~\cite{Zhou14}.  For the networks trained with only RGB data, we follow~\cite{Song15,SGupta14}, which directly use them to extract feature on the depth representation. Then the features extracted from both RGB and depth are concatenated together as inputs for SVM classifier. Note that all models are not fine-tuned on the dataset. As Fig.~\ref{fig:test}(c) shows, our model is $8.2\%$ better than DCGAN and $3.7\%$ away from the Places-AlexNet.

\textbf{Object Detection.} In this task, we perform RGBD object detection on the NYUv2 dataset. We follow the Fast-RCNN pipeline~\cite{girshickICCV15fastrcnn} and use the code and parameter settings provided in~\cite{SGupta16}. In our case, we use surface normal to represent the depth. To apply our model for the detection task, we stacked two fully connected layer ($4096$-d) on top of the last convolutional layer and fine-tune the network end-to-end. We compare against four baselines: network with the same architecture trained from scratch, network pre-trained with DCGAN, DCGANv2, and ImageNet pre-trained AlexNet. For networks pre-trained on only RGB data, we fine-tune them on both the RGB and surface normal inputs separately and average the detection results during testing as~\cite{SGupta16}. We apply Batch Normalization~\cite{BN15} except for ImageNet pre-trained AlexNet. We show the results in Table~\ref{tab:det}. Our approach has $1.5\%$ improvement compared to the model trained from scratch.

\renewcommand{\arraystretch}{1}
\begin{table}[t]
\begin{center}
\setlength{\tabcolsep}{1pt}
\scalebox{0.65}{
\begin{tabular}{c|c|c|c|c|c|c|c|c|c|c|c|c|c|c|c|c|c|c|c|c}
                & mean & bath & bed & book & box & chair & count- & desk & door & dress- & garba- & lamp & monit- & night & pillow & sink & sofa & table & tele & toilet \\
                & & tub & & shelf & & & -er & & & -er & -ge bin & & -or & stand & & & & & vision & \\ \hline

  Ours         & 32.4 & \textbf{44.0} & 67.7 & 28.4 & 1.6 & 34.2 & 43.9 & 10.0 & 17.3 & 33.9 & 22.6 & 28.1 & 24.8 & 41.7 & 31.3 & 33.1 & 50.2 & 21.9 & 25.1 & 54.9 \\
  Scratch         & 30.9 & 35.6 & 67.7 & 23.1 & 2.1 & 33.1 & 40.5 & 10.1 & 15.2 & 31.2 & 19.4 & 26.8 & 29.1 & 39.9 & 30.5 & 36.6 & 43.8 & 20.4 & 29.5 & 52.8 \\
  DCGAN         & 30.4 & 38.9 & 67.6 & 26.3 & \textbf{2.9} & 32.5 & 39.1  & 10.6 & 16.9 & 23.6 & 23.0 & 26.5 & 25.1 & 44.5 & 29.6 & \textbf{37.0} & 45.2 & 21.0 & 28.5 & 38.4 \\
  DCGANv2         & 31.1 & 35.3 & 69.0 & 21.5 & 2.0 & 32.6 & 36.4  & 9.8 & 14.4 & 30.8 & 25.4 & 29.2 & 27.3 & 39.6 & 32.2 & 34.6 & 47.9 & 21.1 & 27.2 & 54.4 \\
  Imagenet         & \textbf{37.6} & 33.1 & \textbf{69.9} & \textbf{39.6} & 2.3 & \textbf{38.1} & \textbf{47.9} & \textbf{16.1} & \textbf{24.6} & \textbf{40.7} & \textbf{26.5} & \textbf{37.8} & \textbf{45.6} & \textbf{49.5} & \textbf{36.1} & 34.5 & \textbf{53.2} & \textbf{25.0} & \textbf{35.3} & \textbf{58.4} \\

\end{tabular}}
\end{center}
\footnotesize{\caption{\label{tab:det}\footnotesize{Detection results on NYU test set.}}
}

\end{table}

\section{Conclusion}
We present a novel Style and Structure GAN which factorizes the image generation process. We show our model is more interpretable and generates more realistic images compared to the baselines. We also show that our method can learn RGBD representations in an unsupervised manner.\\

\noindent \scriptsize{\textbf{Acknowledgement}: This work was supported by ONR MURI N000141010934, ONR MURI N000141612007 and gift from Google. The authors would also like to thank David Fouhey and Kenneth Marino for many helpful discussions.}

\clearpage

\bibliographystyle{splncs}
\bibliography{local}

\newpage
\FloatBarrier

\section{Supplementary Material: Generated Normals and Images from ${\text{S}^2}$-GAN}

\begin{tabular}{c}
\hspace{-0.3in} \includegraphics[width=1.1\textwidth]{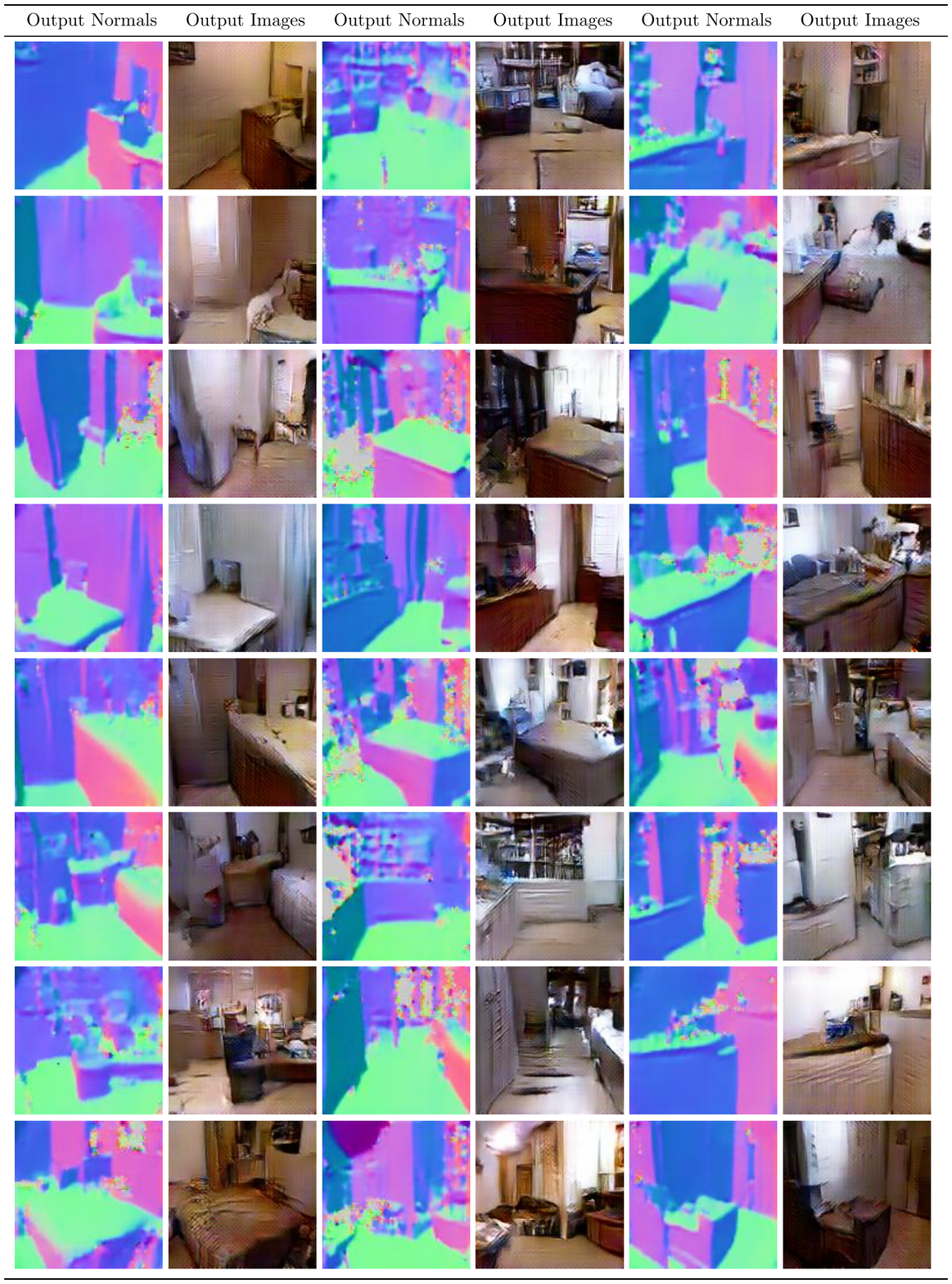}
\end{tabular}

\hspace{-0.5in}
\begin{tabular}{c}
\includegraphics[width=1.1\textwidth]{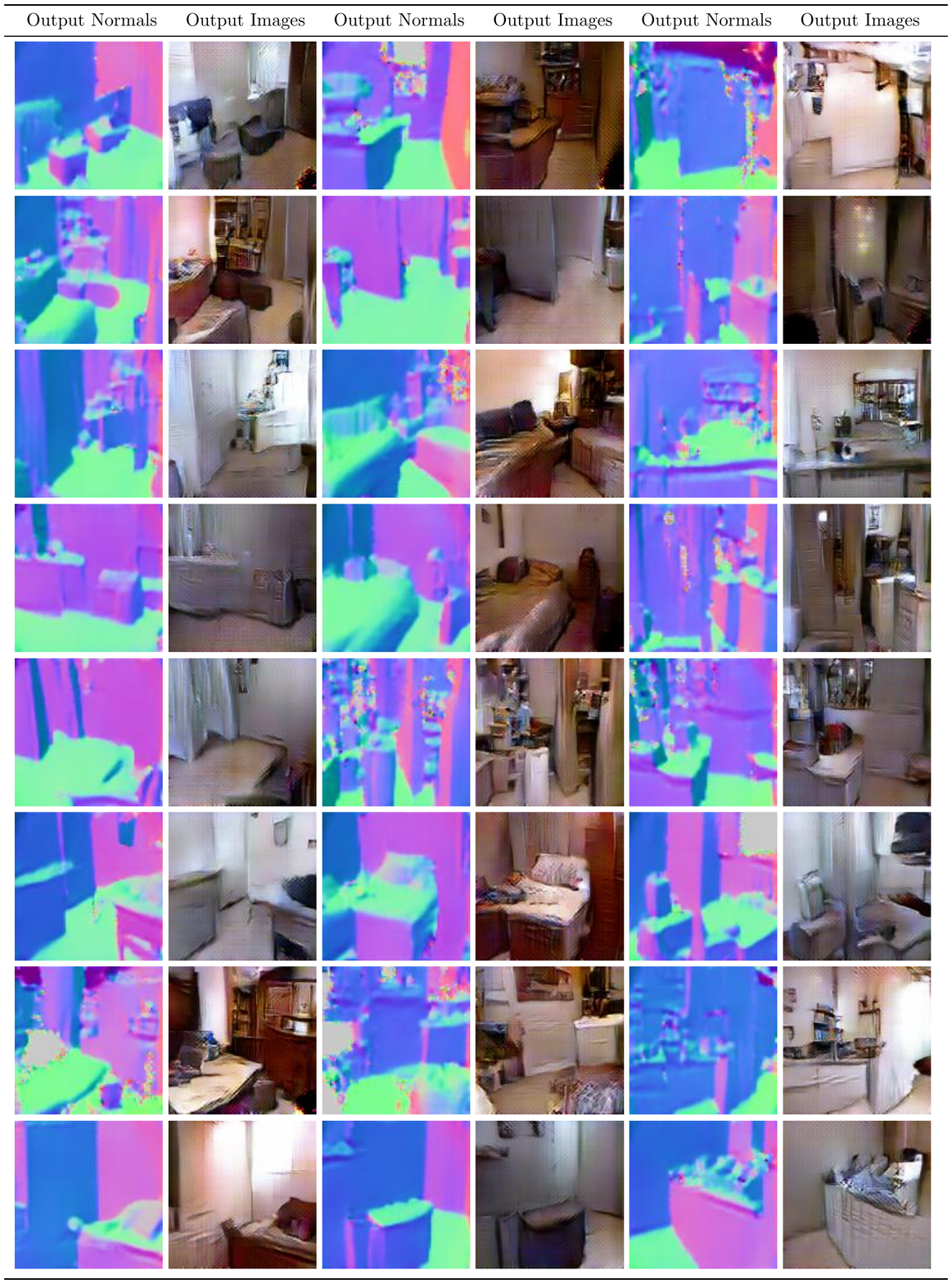}
\end{tabular}

\begin{tabular}{c}
\hspace{-0.5in} \includegraphics[width=1.1\textwidth]{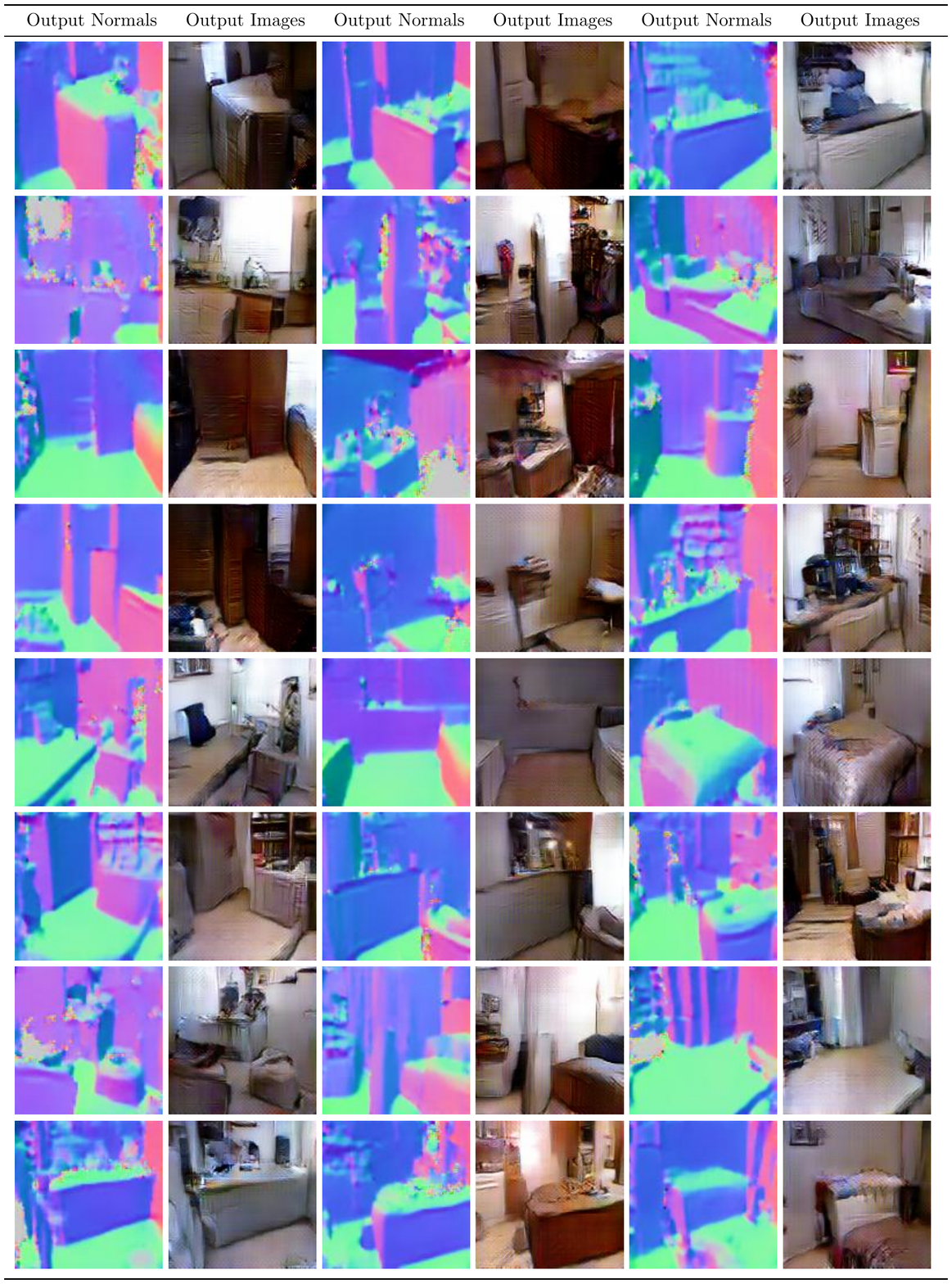}
\end{tabular}

\begin{tabular}{c}
\hspace{-0.5in} \includegraphics[width=1.1\textwidth]{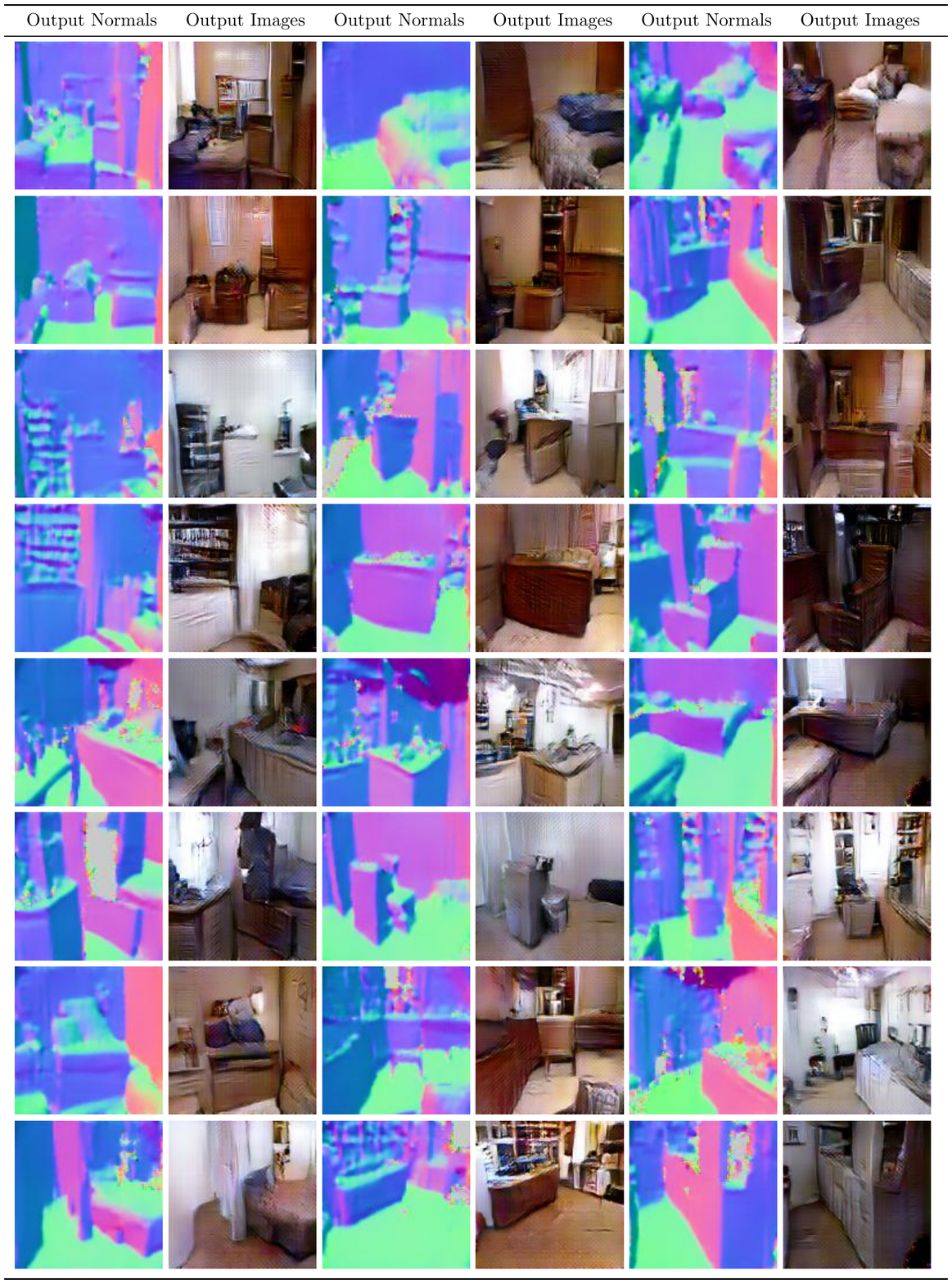}
\end{tabular}

\end{document}